%% file: cameraready_main.tex
\pdfoutput=1
\documentclass[10pt,twocolumn,letterpaper]{article}

\usepackage{cvpr}              % To produce the CAMERA-READY version

\usepackage[accsupp]{axessibility}  % Improves PDF readability for those with disabilities.

\usepackage{graphicx}
\usepackage{amsmath}
\usepackage{amssymb}
\usepackage{booktabs}
\usepackage{enumerate}
\usepackage{multirow}
\usepackage{float}
\usepackage[dvipsnames]{xcolor}

\usepackage{times}
\usepackage{epsfig}
\usepackage{graphicx}
\usepackage{amsmath}
\usepackage{amssymb}
\usepackage{enumerate}
\usepackage{booktabs}
\usepackage{multirow}
\usepackage[dvipsnames]{xcolor}

\usepackage{times}
\usepackage{epsfig}
\usepackage{graphicx}
\usepackage{amsmath}
\usepackage{amssymb}
\usepackage{tabulary}
\usepackage{multirow}
\usepackage{silence}
\usepackage{caption}
\usepackage{xfrac}
\usepackage{float}
\usepackage[pagebackref=true,breaklinks=true,letterpaper=true,colorlinks,bookmarks=false]{hyperref}

\makeatletter
\renewcommand\paragraph{\@startsection{paragraph}{4}{\z@}%
                                    {0.5ex \@plus1ex \@minus.1ex}%
                                    {-1em}%
                                    {\normalfont\normalsize\bfseries}}
\makeatother

\usepackage[capitalize]{cleveref}
\crefname{section}{Sec.}{Secs.}
\Crefname{section}{Section}{Sections}
\Crefname{table}{Table}{Tables}
\crefname{table}{Tab.}{Tabs.}

\def\cvprPaperID{988}
\def\confName{CVPR}
\def\confYear{2022}

\begin{document}

\title{\emph{CaDeX}: Learning \emph{Ca}nonical \emph{De}formation \emph{C}oordinate \emph{S}pace 
for Dynamic Surface Representation 
via Neural Homeomorphism}

\author{Jiahui Lei \quad
Kostas Daniilidis \quad
\\
\vspace{2mm}
University of Pennsylvania \\
\small{\url{https://www.cis.upenn.edu/~leijh/projects/cadex}} \\
}

\twocolumn[\maketitle\vspace{-0.5em}\input{./main_text/0teaser.tex}\bigbreak]

\input{./main_text/0abs.tex}
\input{./main_text/1intro.tex}
\input{./main_text/2related.tex}
\input{./main_text/3method}
\input{./main_text/4exp.tex}
\input{./main_text/5dis.tex}
\input{./main_text/6conclusion.tex}

\newpage
{\small
\bibliographystyle{ieee_fullname}
\bibliography{reference.bib}
}

\end{document}

%% file: main_text/0teaser.tex
\begin{center}
\vspace{-3em}
\includegraphics[width=1\linewidth]{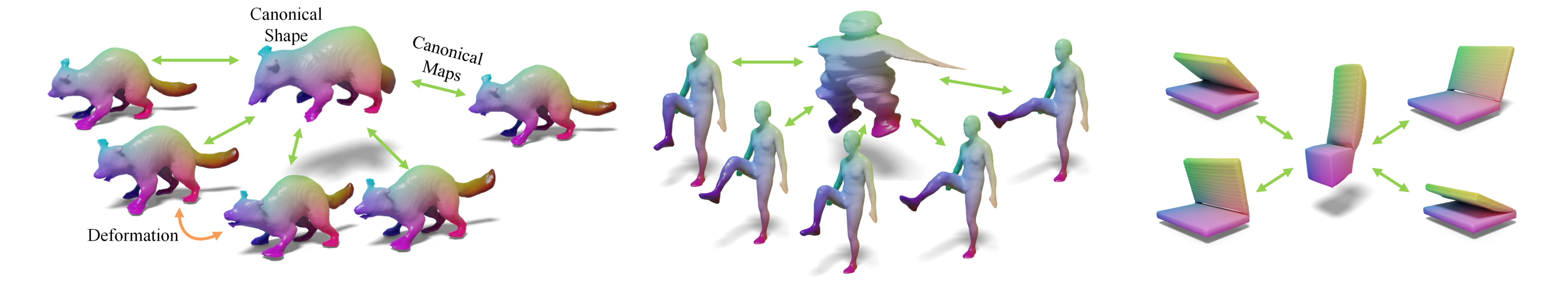}
\end{center}\vspace{-2em}
\captionof{figure}{We model the deformable surface through a learned \textbf{canonical shape} (the middle shape) and factorize the deformation (orange arrow) by learnable continuous bijective \textbf{canonical maps} (green \textbf{bidirectional} arrows) that provide the cycle consistency and topology preservation. Visual results are from the test set on dynamic animals, human bodies and articulated objects. 
}
\label{fig:teaser}

%% file: main_text/0abs.tex
\begin{abstract}
While neural representations for static 3D shapes are widely studied, representations for deformable surfaces are limited to be template-dependent or to lack efficiency. We introduce Canonical Deformation Coordinate Space (CaDeX), a  unified representation of both shape and nonrigid motion.  Our key insight is the factorization of the deformation between frames by continuous bijective canonical maps (homeomorphisms) and their inverses that go through a learned canonical shape. Our novel deformation 
representation and its implementation are simple, efficient, and guarantee cycle consistency, topology preservation, and, if needed, volume conservation. Our modelling of the learned canonical shapes provides a flexible and stable space for shape prior learning. We demonstrate state-of-the-art performance in modelling a wide range of deformable geometries: human bodies, animal bodies, and articulated objects.
\end{abstract}

%% file: main_text/1intro.tex
\vspace{-1.5em}
\section{Introduction}

Humans perceive, interact, and learn in a continuously changing real world. One of our key perceptual capabilities is the modeling of a dynamic 3D world.
Such geometric intelligence requires sufficiently general neural representations that can model different \textbf{dynamic} geometries in 4D sequences to facilitate solving robotics~\cite{xu2020learning}, computer vision~\cite{lai2021video}, and graphics~\cite{dnerf} tasks. 
Unlike the widely studied 3D neural representations, a dynamic representation has to be able to associate (for example, finding correspondence) and aggregate (for example, reconstruction and texturing) information across the deformation states of the world. Directly extending a successful static 3D representation (for example,~\cite{onet}) to each deformed frame leads to low efficiency~\cite{oflow}, and the inability to model the  information flow across frames, which is critical when solving ill-posed problems as in~\cite{dnerf}. 
Our desired dynamic representation needs to simultaneously represent a global surface (\textbf{canonical/reference shape}) across all frames and the \textbf{consistent deformation} (correspondence/flow/motion) between any frame pair (Fig.~\ref{fig:teaser}), so that we can recover the dynamic geometry by reconstructing only one reference surface and generating the rest of the deformed surfaces by using the consistent deformation representation as well as associate and aggregate information across frames (Fig.~\ref{fig:formulation}A). 

The majority of dynamic representations that satisfy the above desired properties are model-based and rely on parametric models for specific categories like human bodies~\cite{anguelov2005scape,SMPL:2015} (Fig.~\ref{fig:formulation}B), faces~\cite{blanz1999morphable,li2017learning}, or hands~\cite{MANO:SIGGRAPHASIA:2017}. 
On the contrary, recent model-free methods like the implicit flow \cite{oflow,lpdc} (Fig.~\ref{fig:formulation}C) apply one universal 4D representation but model the canonical shape in an ad hoc chosen frame~\cite{lpdc,oflow} that complicates the shape prior.  Alternatively, the choice of an approximate mean/neutral shape~\cite{yenamandra2021i3dmm} as the canonical shape can limit the shape expressibility.  Modeling of the deformation is done by either  MLPs~\cite{lpdc,yenamandra2021i3dmm} that ignore the real world deformation properties, or by ODEs~\cite{oflow} that are inefficient for space deformation, or by an optimized embedded graph~\cite{neuraldeformgraph} or Atlas~\cite{bednarik2021temporally} that are sequence specific.

In this work, we introduce a novel and general  architecture and representation that enable a competitive reconstruction of every frame and the recovery of consistent correspondence across frames. 
Our approach is rooted in the factorization of deformation (Sec.~\ref{sec:cadex_math}). If we assume that the topology does not change during deformation, all deformed surfaces of one instance can be regarded as equivalent through continuous bijective mappings (homeomorphisms).
This allows us to factorize the deformation between two frames by the composition of two continuous invertible functions such that one maps the source frame into a common 3D Canonical Deformation Coordinate Space (\textbf{CaDeX}) while another maps it back to the destination frame. Such a factorization and its implementation (Sec.~\ref{sec:canonical_map}) is novel, simple, and efficient (compared to ODEs~\cite{oflow}) while it guarantees cycle consistency, topology preservation, and, if necessary, volume conservation (Sec.\ref{sec:cadex_properties}). 
The {\bf canonical shape} embedded in the CaDeX can be regarded as the representative element, while the associated invertible mappings that transform between deformed frames and the CaDeX are the \textbf{canonical maps}. 
Therefore, we model the reference surface directly in the CaDeX via an implicit field~\cite{onet} (Sec.~\ref{sec:canonical_shape}), which can be optimized
together with the canonical maps during training.

In summary, our main \textbf{contributions} are: (1) A novel general representation and architecture for dynamic surfaces that jointly solve the canonical shape and consistent deformation problems. 
(2) Learnable continuous bijective canonical maps and canonical shapes that jointly factorize the shape deformation, and are novel, simple, efficient, and guarantee cycle consistency and topology preservation.
(3) A novel solution to the dynamic surface reconstruction and correspondence tasks given sparse point clouds or depth views based on the proposed representation.
(4) We demonstrate state-of-the-art performance on modelling different deformable categories: Human bodies~\cite{dfaust:CVPR:2017}, Animals~\cite{dt4d} and Articulated Objects~\cite{wang2019shape2motion}.

%% file: main_text/2related.tex
\section{Related Work}
\label{sec:related}
Proposed neural representations for static 3D geometry~\cite{onet,deepsdf,imnet,genova2020local,genova2019deep,peng2020convolutional,jiang2020local,oechsle2021unisurf,mildenhall2020nerf,yang2019pointflow,atlasnet,chibane2020implicit} are promising, but most of them do not involve modeling of deformations. A few recent approaches represent or process 3D shapes via deformation~\cite{shapeflow,neuralmeshflow,meshode,deformedif,zheng2021deep}, but they focus on static 3D shape collections that do not meet the requirements (e.g, efficiency) for processing 4D data.
We will focus  our related work on dynamic representations of deformable geometry.

\begin{figure}[t]
    \centering
   \includegraphics[width=\columnwidth]{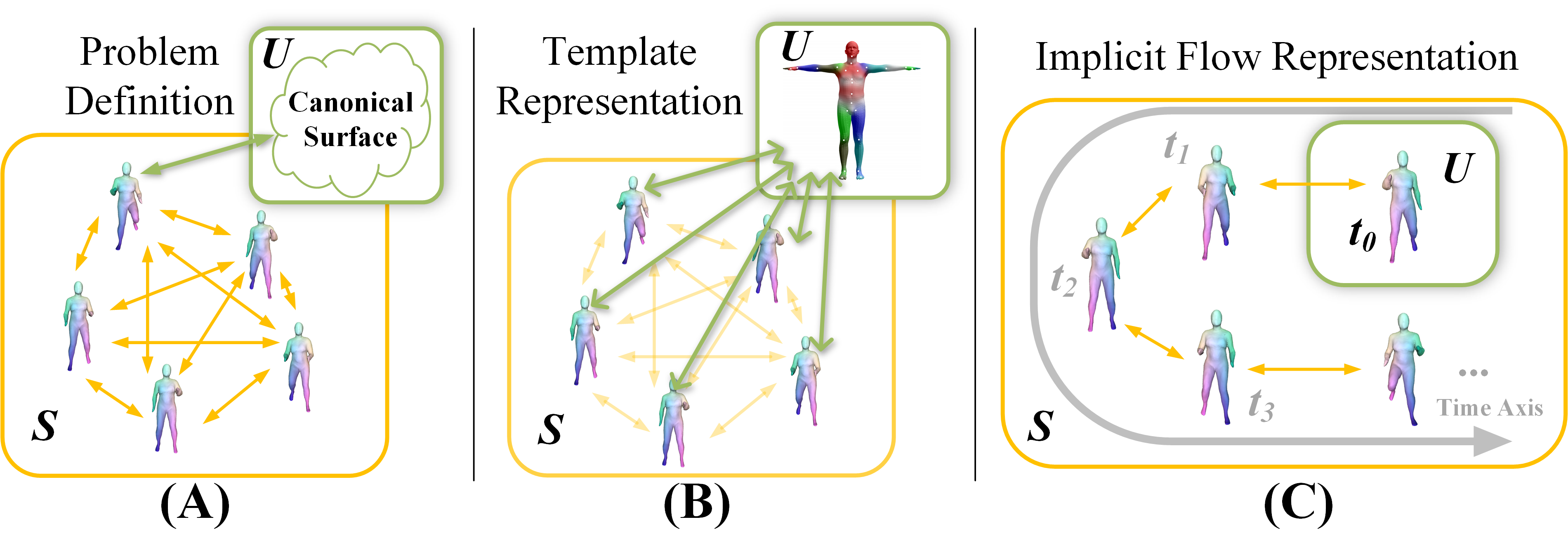}
    \vspace{-2em}
    \caption{
\textbf{(A) Problem definition}:
A list, or a set of deformed surfaces $\mathcal{S}=\{S_i\}$ of one instance should be represented by 1.) one canonical 3D surface $U$ (in the green box) and 2.) the consistent deformation between surfaces (yellow arrows); \textbf{(B) Model-based methods}: $S_i$ is obtained through the skinning function (green arrows) from the template mesh~\cite{SMPL:2015}; \textbf{(C) Implicit-flow methods}: the first frame serves as the reference shape and the deformation is modeled by Neural-ODEs~\cite{oflow} or MLPs~\cite{lpdc}.
    }
    \label{fig:formulation}
\end{figure}

\noindent \textbf{Model-Based Dynamic Representation}:
Many successful 3D parametric models for specific shape  categories have been introduced, for example, the morphable model~\cite{blanz1999morphable} and FLAME~\cite{li2017learning} for faces, 
SCAPE~\cite{anguelov2005scape} and SMPL~\cite{SMPL:2015} for human bodies,  and MANO~\cite{MANO:SIGGRAPHASIA:2017} for hands .
These model-based representations (Fig.~\ref{fig:formulation}B)  
suffer from limited expressivity, which can be mitigated by neural networks.
Networks can express detailed template shapes based on template meshes ~\cite{cape,scanimate,metaavatar,npms} or skeletons~\cite{nasa,neuralgif,karunratanakul2021skeleton}, and can learn more detailed skinning functions~\cite{snarf,scanimate,metaavatar} or forward deformations~\cite{npms}. 
However, they rely on the strong assumption of canonicalization through the pose, skeleton, and template mesh, which makes them limited to specific categories, and insufficient for modeling the rich dynamic 3D world. 
Our method does not rely on any hard-wired template mesh or skeleton, and the same architecture is universal for all shape categories.

\noindent \textbf{Model-Free Dynamic Representation}:
Recent works ~\cite{oflow,lpdc,lcr,neuraldeformgraph} extend the success of static 3D representations~\cite{onet,deepsdf,imnet} to 4D by modeling the deformation between frames. 
Fig.~\ref{fig:formulation}C illustrates how the two closest works to ours, O-Flow~\cite{oflow} and LPDC~\cite{lpdc}, are related to our problem formulation. 
First, our method differs from O-Flow~\cite{oflow} and LPDC~\cite{lpdc} in the representation of the space deformation. We represent the deformation through a novel canonical map factorization that is efficient and guarantees real world properties based on conditional neural homeomorphisms~\cite{realnvp,nice}, while O-Flow~\cite{oflow} uses a Neural-ODE~\cite{neuralode} that also guarantees the production of a well-behaved deformation (see ~\cite{neuralmeshflow} for details) but with higher computational complexity than ours.
LPDC~\cite{lpdc} replaces the Neural-ODE~\cite{neuralode} by a Multilayer Perceptron (MLP)
to learn correspondences in parallel. However, the MLP deformation~\cite{atlasnet,lpdc,dnerf,deformablenerf} has difficulty to model a homeomorphism or express real world deformation properties.
Note that both O-Flow~\cite{oflow} and LPDC~\cite{lpdc} compute the reference surface in the first frame, which turns out to be a random choice, since the shape can be in an arbitrary deformation state in the first frame. 
Our reference shape is modeled in the learned canonical space induced by the canonical map, which is more stable and can be optimized (Fig.~\ref{fig:teaser}).

I3DMM~\cite{yenamandra2021i3dmm} learns a near neutral/mean canonical template from human head scans, which limits its expressive ability. CASPR \cite{caspr} and Garment Nets~\cite{chi2021garmentnets} learn a canonicalization of deformable objects, but rely on the ground truth canonical coordinate supervision, which is often inaccessible.
Other neural dynamic representations include the learned embedded graph \cite{neuraldeformgraph} first proposed in 
\cite{sumner2007embedded} and parametric atlases \cite{bednarik2021temporally}.
Beyond 4D data, A-SDF~\cite{asdf} models the general articulated objects with a specially designed disentanglement network, but it cannot model  correspondence. Instead, our method achieves stronger disentanglement by explicitly modeling the deformation.

\noindent \textbf{Invertible Networks for 3D representation}:
Many works~\cite{germain2015made,nice,papamakarios2017masked,realnvp,kingma2018glow,neuralode,behrmann2019invertible} have been proposed to construct invertible networks for generative models. In 3D deep learning,
Neural-ODE \cite{neuralode} is widely used as a good model of deformation~\cite{shapeflow,neuralmeshflow,meshode,oflow} or transformation of point cloud~\cite{yang2019pointflow}. ShapeFlow~\cite{shapeflow} learns a ``Hub-and-spoke" surface deformation for 3D shape collection via ODEs, but is inefficient when applied to the 4D data since every frame needs to be lifted to the ``hub" through integration.
Besides ODEs, I-ResNet~\cite{behrmann2019invertible} is used in ~\cite{yang2201GPNF} to build invertible deformation for shape editing.
Our method is inspired by Neural-Parts~\cite{neuralparts} where Real-NVP~\cite{realnvp} is used to model the deformation from a sphere primitive to a local part. While we also use Real-NVP~\cite{realnvp} for its simplicity and efficiency, we have two distinct differences compared to \cite{neuralparts}: Our canonical shape is a learned implicit surface instead of a fixed sphere that can only capture local parts; we use the inverse of the Real-NVP to close the factorization cycle, while \cite{neuralparts} uses the inverse in a complementary training path.

%% file: main_text/3method.tex
\section{Method}

\begin{figure*}[ht!]
    \centering
  \includegraphics[width=\textwidth]{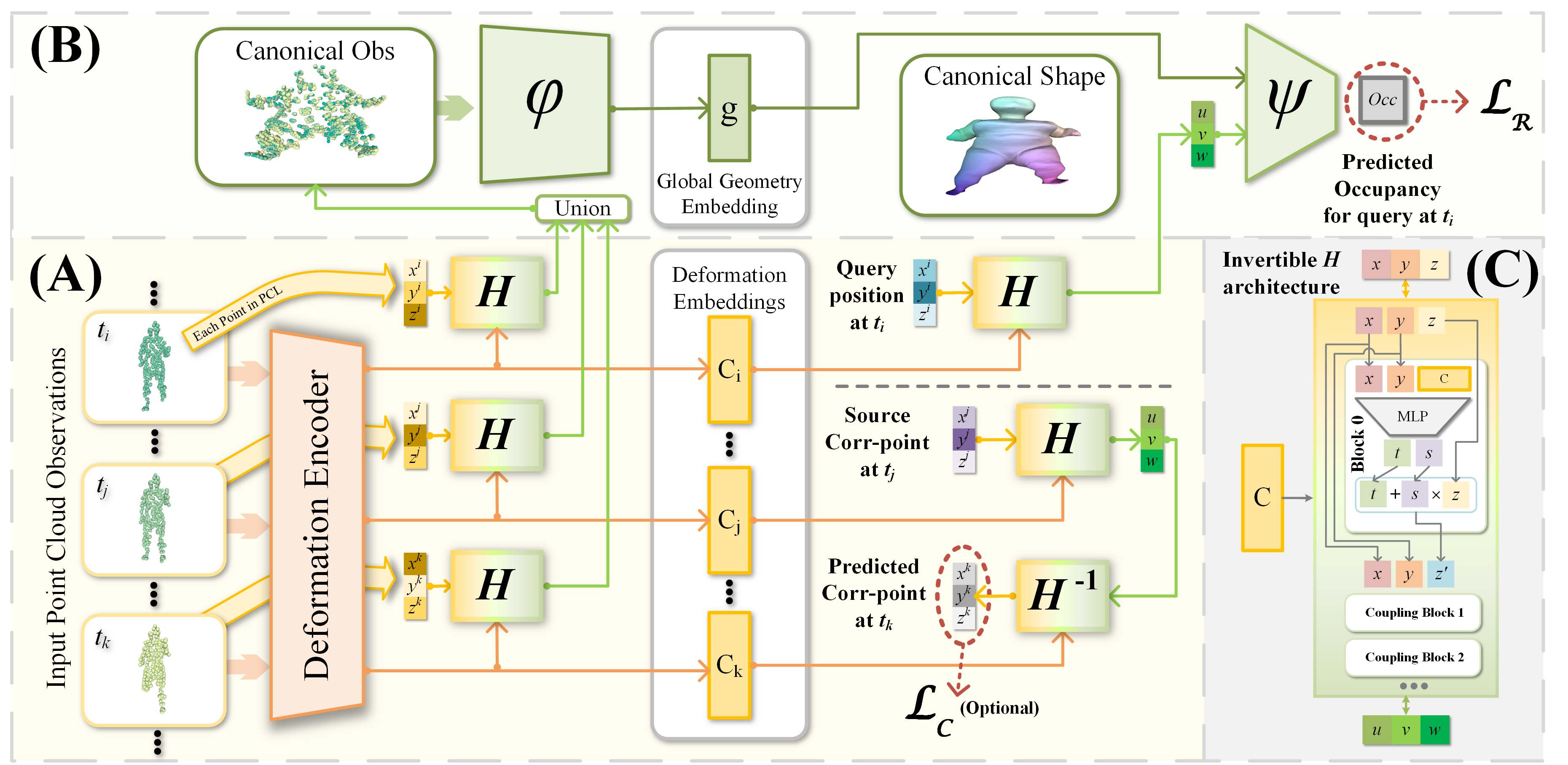}
    \caption{
    \textbf{(A) Canonical Map} (Sec.~\ref{sec:canonical_map}): A sequence (or a set) of input point clouds is first sent to the Deformation Encoder generating deformation embeddings $c_i$ for each frame.
    % (Sec.~\ref{sec:deformation_encoder}). 
    Then the canonical map $H$ can transform any coordinate (e.g. a yellow point in the point cloud, a blue query position for implicit field or a purple source point for correspondence) from any deformed frame to the canonical coordinate via $H$ conditioning on the corresponding deformation embedding. The correspondence prediction (right bottom) can be obtained by directly mapping back the canonical coordinate through $H^{-1}$. 
    \textbf{(B) Canonical Shape Encoder-Decoder} (Sec.~\ref{sec:canonical_shape}): All input multi-frame point clouds are first transformed to the canonical space via $H$ and are directly unioned to aggregate a canonical observation. The global geometry embedding $g$ (unique across frames) is encoded via a PointNet~\cite{qi2017pointnet} $\phi$, and the occupancy value for the canonical coordinate of a query position at $t_i$ (blue point) is predicted through a standard OccNet~\cite{onet} $\psi$.
    During training, the occupancy is supervised by $\mathcal L_R$, and the correspondence can be optionally supervised by $\mathcal L_C$ (Sec.~\ref{sec:losses}).
    \textbf{(C) The Real-NVP~\cite{realnvp} invertible architecture of $H$} (Sec.~\ref{sec:canonical_map}).
    }
    \label{fig:arch}
\end{figure*}

\noindent As in Fig.~\ref{fig:formulation}A, given a sequence\footnote{Or a set, but for conciseness, we will only refer to the sequence.} of point cloud observations of one deforming instance, our goal is to reconstruct a sequence of surfaces.
Instead of directly solving a correspondence map between two frames during reconstruction, we propose an architecture (Fig.~\ref{fig:arch}) where the interframe correspondence is computed via a pivot {\bf canonical shape}. We will call the map between a surface in any deformed frame and the canonical shape a  {\bf canonical map}.

\subsection{CaDeX and Canonical Map}
\label{sec:cadex_math}
Let us denote $[x^{i},y^{i},z^{i}]\in \mathbb R^3$ as the 3D coordinates of the input 3D space\footnote{The superscript $i$ refers to the time index.}, in which a deformed surface $S_i$ at time $t_i$ is embedded. 
Consider a continuous bijective mapping (homeomorphism) $\mathcal H_i: \mathbb R^3 \mapsto \mathbb R^3$ at time $t_i$ that maps each deformed coordinate to its global (shared over different time frames) 3D coordinate $[u,v,w] = \mathcal H_i([x^{i},y^{i},z^{i}])$. 
Note that $[u,v,w]$ has no index of time and can be seen as a globally consistent indicator of each correspondence trajectory across time in the input 3D space. 
Hence, we name $[u,v,w]$ the {\bf canonical deformation coordinates} of the position $[x^{i},y^{i},z^{i}]$ at time $t_i$ and call the $uvw$ 3D space the \textbf{Canonical Deformation Coordinate Space} (CaDeX) of the sequence $\mathcal S=\{S_i\}$. The homeomorphisms $\mathcal H_t$ that transform $[x^{t},y^{t},z^{t}]$ to $[u,v,w]$ are called \textbf{canonical maps}.
Since CaDeX is globally shared across time, we model the \textbf{canonical shape} (surface) $U$ directly in CaDeX instead of selecting an input frame as is the case in~\cite{oflow,lpdc}.
Taking advantage of neural fields~\cite{xie2021neural}, we model $U$ as a level set of an occupancy field~\cite{onet}:
\begin{equation}
    U=\left\{\, [u,v,w]\,|\,\text{OccField}([u,v,w])=l\, \right\},
    \label{eq:canonical_shape}
\end{equation}
where $l$ is the surface level. Using the inverse of each canonical map $\mathcal H_i$ at time $t_i$, we can directly obtain each deformed surface at time $t_i$  in the input 3D space as:
\begin{equation}
 S_i=\left\{\, \mathcal H^{-1}_i([u,v,w]) \,|\, \, \forall [u,v,w] \in U \, \right\}.\label{eq:deformed_shape}
\end{equation}
The correspondence/deformation $\mathcal F_{ij}$ that associates any coordinate (for both surface and non-surface points) from the 3D space at time $t_i$ to the 3D space at time $t_j$ can be factorized by the canonical maps as:
\begin{equation}
    [x^{j},y^{j},z^{j}] = \mathcal F_{ij}([x^{i},y^{i},z^{i}]) = \mathcal H_j^{-1}\circ \mathcal H_i ([x^{i},y^{i},z^{i}]). \label{eq:flow}
\end{equation}
Note that $\mathcal H_t$ must be invertible; otherwise, the above deformation function cannot be defined. By now, any surface that is topologically isomorphic to the deformable instance satisfies the above definitions, leading to infinitely many valid canonical shapes and maps. In the following, we will optimize the canonical shapes and maps predicted by the architecture in Fig.~\ref{fig:arch} subject to the priors from the dataset.

\subsection{Canonical Map Implementation}
\label{sec:canonical_map}
\paragraph{Neural Homeomorphism}
\label{sec:inn}
One key technique of our implementation is an efficient way to parameterize and learn the homeomorphism between coordinate spaces. Unfortunately, the widely used Neural-ODEs~\cite{neuralode} do not meet our efficiency requirements since a full integration would have to be applied to every frame.
Inspired by~\cite{neuralparts}, we utilize the Conditional Real-NVP~\cite{realnvp} (Real-valued Non-Volume Preserving) or the NICE~\cite{nice} (Nonlinear Independent Component Estimation) normalizing flow implementations to learn the homeomorphism. Taking the more general NVP~\cite{realnvp} as an example (Fig.~\ref{fig:arch}-C), with the network being a stack of Coupling Blocks \cite{realnvp}, we apply NVP to 3D coordinates. During initialization, each block is randomly assigned an input split pattern; for example, a block always splits $[x,y,z]$ to $[x,y]$ and $[z]$. Given a condition latent code $c$, each block takes in 3D coordinates $[x,y,z]$ and outputs the transformed coordinates $[x',y',z']$ by changing one part based on the other part in the input coordinate split:
\begin{equation}
\left[x',y',z'\right] = \left[x, y, z\exp(s_\theta(x,y|c))+t_\theta(x,y|c)\right]
\end{equation}
where $s_{\theta}(\cdot |c) : \mathbb{R}^2 \mapsto \mathbb{R}$ and $t_{\theta}(\cdot |c) : \mathbb{R}^2 \mapsto \mathbb{R}$ are scale and translation predicted by any network conditioned on $c$. Such a block models a bijection since the inverse can be immediately derived as:
\begin{equation}
\left[x,y,z\right] = \left[x', y', \frac{z'-t_\theta(x',y'|c)}{\exp(s_\theta(x',y'|c))}\right].
\end{equation}
Therefore, the whole stack of blocks is invertible. If the activation functions in each block are continuous, then the whole network models a homeomorphism. NICE~\cite{nice} is simply removing the scale freedom from the NVP block, i.e: $s_{\theta}(\cdot |c)\equiv0$. Note that the inverse of NVP and NICE is as simple as the forward, which induces our desired efficiency and simplicity, and enables using the definition in Eq.\ref{eq:flow}.

\paragraph{$\mathcal H$ architecture}
\label{sec:h_archi}
Note that in Eq.~\ref{eq:deformed_shape}, each deformed surface $S_i$ has a different canonical map $\mathcal H_i$ that associates $S_i$ with $U$. We implement them with the conditional real-NVP or NICE, denoted by $H$ (noncalligraphic). 
Given the vector $c_{i}$ that encodes the deformation information at time $t_i$ such that $\mathcal H_i(\cdot) \equiv H(\cdot\,;\, c_{i})$, where the network $H$ is shared across different time frames.
The canonical deformation coordinates are predicted as (Fig.~\ref{fig:arch}-A, boxes marked with $H$):
\begin{equation}
    [u,v,w] = H([x^{i},y^{i},z^{i}]\,;\, c_{i}).
    \label{eq:cdc}
\end{equation}
Note that on the right side of Eq.~\ref{eq:cdc}, the input coordinates and the deformation embedding have the index $t_i$ since they come from each deformed frame. However, after application of the canonical map, the coordinates on the left side are independent of the index because there is only one global CaDeX for this sequence. 
Finally, the correspondence/deformation between two deformed frames (Eq.~\ref{eq:flow}) can be implemented as:
\begin{equation}
    [\hat{x}^{j},\hat{y}^{j},\hat{z}^{j}] = H^{-1}\left(
        H([x^{i},y^{i},z^{i}]\,; \,c_{i})
        \,;\, c_{j}\right)
    \label{eq:corr_pred}
\end{equation}
where $[\hat{x}^{j},\hat{y}^{j},\hat{z}^{j}]$ is the mapped position at time $t_j$ of the original position $[x^{i},y^{i},z^{i}]$ at time $t_i$.
Regarding the choice of $H$, Real-NVP~\cite{realnvp} can provide more flexible deformation since it has one more degree of freedom (scale); NICE~\cite{nice} guarantees volume conservation (Sec.~\ref{sec:cadex_properties}) that results in a more regularized deformation.

\paragraph{Deformation Encoder}
\label{sec:deformation_encoder}
To obtain the per-frame deformation embedding $c_i$ that is used as the condition of $H$,
we demonstrate two kinds of inputs and three encoder types (Fig.~\ref{fig:arch}-A, orange box). 
One direct approach is to employ a PointNet that summarizes the deformation code separately per frame (\emph{PF}). If the input is a sequence of point clouds, we can alternatively use the ST-PointNet variant proposed in \cite{lpdc} to get the deformation code (\emph{ST}). The \emph{ST} encoder processes the 4D coordinates and applies the pooling spatially and temporally. 
If the input is a set without order, we develop a 2-phase PointNet to obtain a global set deformation code (\emph{SET}), and then use a 1-D code query network to output the deformation embedding $c_{i}$ taking the query articulation angle and the global deformation code as input.
Since these are not our main contributions, we refer the reader to the supplementary for details on these encoders.

\subsection{Properties of the Canonical Map}
\label{sec:cadex_properties}
The novel factorization and its implementation induce the following desired properties of real world deformation:

\noindent \textbf{\emph{Cycle consistency}}:
The deformation/correspondence between  deformed frames predicted by our factorization (Eq.~\ref{eq:flow},~\ref{eq:corr_pred}) is cycle consistent (path-invariant). The reason is that every canonical map maps any deformed frame in the sequence (or set) to the global CaDeX of this sequence (or set), and the canonical maps are invertible:
\begin{equation}
\begin{aligned}
    \mathcal F_{jk}\circ \mathcal F_{ij} &= \mathcal H_k^{-1}\circ \mathcal H_j \circ \mathcal H_j^{-1}\circ \mathcal H_i
    = \mathcal H_k^{-1}\circ \mathcal H_i = \mathcal F_{ik}.
\end{aligned}
\end{equation}
    
\noindent \textbf{\emph{Topology preserving deformation}}: 
Since our factorization (Eq.~\ref{eq:flow},~\ref{eq:corr_pred}) is a composition of two homeomorphisms, the induced deformation function is thus a homeomorphism as well, and therefore  never changes the surface topology. 

\noindent \textbf{\emph{Volume conservation (NICE)}}: If $H$ is implemented by NICE~\cite{nice}, then the predicted deformation preserves the volume of the geometry, which can be proved by the fact that the determinant of the Jacobian of every coupling block in NICE~\cite{nice} is $1$ (see Supp. for more details).
    
\noindent \textbf{\emph{Continuous deformation if $c$ is continuous}}:  
Some applications require the sequence $\mathcal S=\{S\}$ to be dense on time axis, for example, modeling continuous deformation across time~\cite{oflow}. In this case, the deformation codes $c$ become a function of time $c(t)$. Since all activation functions we are using in the canonical map 
are continuous, it is obvious that if $c(t)$ is continuous, then the predicted deformation in Eq.~\ref{eq:corr_pred} must be continuous across $t$.

\subsection{Representing Canonical Shape}
\label{sec:canonical_shape}
\paragraph{Geometry Encoder}
\label{sec:g_encoder}
We represent the canonical shape in the CaDeX by a standard Occupancy Network~\cite{onet}. 
The canonical map brings additional benefits for encoding the global geometry embedding (Fig.~\ref{fig:arch}-B).
Denote the observed point cloud at time $t_i$ as $X_{i} = \{\,[x_j^{i}, y_j^{i}, z_j^{i}] \,|\,j=0,1,\ldots,N_i\,\}$\footnote{Note here the superscripts are still the index of the time, the subscripts are the index of the points in the cloud.}, where $[x_j^{i}, y_j^{i}, z_j^{i}]$ is the 3D coordinate of each point in the point cloud.
The observations from different $t_i$'s are partial, noisy, and not aligned.
We overcome such irregularity by using the same canonical map (Sec.\ref{sec:canonical_map}) to obtain a canonical aggregated observation. Given the deformation embedding $c_{i}$ per-frame, the canonical observations are merged via set union as:
\begin{equation}
    \bar{X} = \bigcup_{t_i}\{\, H([x_j^{i}, y_j^{i}, z_j^{i}]\,;\, c_{i})\,|\, \forall [x_j^{i}, y_j^{i}, z_j^{i}] \in X_i \,\}.
\end{equation}
The global geometry embedding $g$ of the sequence $\mathcal S$ is encoded by a PointNet $\phi$: $g = \phi (\bar{X})$.

\paragraph{Geometry Decoder}
\label{sec:g_decoder}
Given the global geometry embedding $g$, we obtain the canonical shape encoded in $g$ via an occupancy network~\cite{onet}  that takes $g$ as well as the query position $[u,v,w]$ in the CaDeX as input, and predicts the occupancy in the CaDeX:  $\hat{o} = \psi([u,v,w];\, g)$, where the decoder $\psi$ is an MLP. 
However, the ground truth $([u,v,w],o^*)$ supervision pair is unavailable in the CaDeX since the canonical shape 
is not known in advance and is learned during training.
Available types of supervision are the query-occupancy pairs $([x^{i}, y^{i}, z^{i}],o^{i*})$ in each deformed coordinate space where the deformed surface $S_i$ is embedded at each time $t_i$. 
Therefore, we predict the occupancy field of any deformed frame through the canonical map via Eq.~\ref{eq:cdc}:
\begin{equation}
    \hat{o} = \psi\left(H([x^{i}, y^{i}, z^{i}]\,;\, c_{i})\,;\, g\right)
    \label{eq:occ_pred}
\end{equation}

\subsection{Losses, Training, Inference}
\label{sec:losses}
Our model is fully differentiable and is trained end-to-end. Following  Eq.~\ref{eq:occ_pred}, the main loss function is the reconstruction loss in each deformed frame:
\begin{equation}
    \mathcal L_{R} = \frac{1}{T}\sum_{i=1}^T{
    \frac{1}{M_i} \sum_{j=1}^{M_i}
    BCE \left[
    \psi\left(H(p_j^{i}; c_{i}); g\right)
    , o_j^{i*}
    \right]
    \label{eq:l_recon}
    }
\end{equation}
where $T$ is the total number of frames that have occupancy field supervision and $M_i$ is the number of queried positions at each frame. We denote by $p_j^{i}$ the $j^{th}$ query position in frame $t_i$, and by $o_j^{i*}$ the corresponding ground truth occupancy state.
Optionally, if the ground truth correspondence pairs are given, we can utilize them as a supervision signal via Eq.~\ref{eq:corr_pred}.  The additional correspondence loss reads:
\begin{equation}
    \mathcal L_{C} = \frac{1}{|\mathcal Q|}\sum_{(p_k^{i},p_k^{j})\in \mathcal Q} {\left\|
    H^{-1}\left(
        H(p_k^{i};c_{i})
        ;c_{j}\right)  - p_k^{j}
    \right\|_l}
    \label{eq:l_corr}
\end{equation}
where $\mathcal Q$ is the set of ground truth correspondence pairs: $p_k^{i}$ is the source position (Fig.~\ref{fig:arch} purple coordinate) in frame $t_i$ and $p_k^{j}$ is the ground truth corresponding position in frame $t_j$; and $k$ is the index of all supervision pairs. We denote by $l$ the order of the error norm. 
Note that the cycle consistency guaranteed by our method (Sec.~\ref{sec:cadex_properties}) does not depend on $\mathcal L_C$.
The overall loss function is $\mathcal L = w_R\mathcal L_R + w_C\mathcal L_C$, where $w_C$ can be zero if no correspondence supervision is provided.
Note that there is no loss directly applied to the 
predicted canonical deformation coordinates $[u,v,w]$. 
This gives the maximum freedom to the canonical shape to form a pattern that helps the prediction accuracy.
All patterns of the canonical shape emerge automatically during training (see the Supplement for an additional discussion).

During training, our model is trained directly from scratch with the mandatory reconstruction loss (Eq.~\ref{eq:l_recon}). For efficiency, at each training iteration, we randomly select a subset of frames in the input sequence and supervise the occupancy prediction. If the ground-truth correspondence supervision is also provided, we predict the corresponding position of surface points in the first frame for every other frame and minimize the correspondence loss in Eq.~\ref{eq:l_corr}.

During inference, our model generates all surfaces of a sequence in parallel after a single marching cubes mesh extraction.
Directly marching the CaDeX is intractable since it is learned. However, by using Eq.~\ref{eq:occ_pred} as a query function, we can extract the mesh $(\mathcal V_0, \mathcal E_0)$ in the first frame, which is equivalent to marching the CaDeX given the canonical map. The equivalent canonical mesh in the CaDeX is $(\mathcal V_c, \mathcal E_c) = (H(\mathcal V_0;c_{0}),\mathcal E_0)$.
Then any mesh in other frames can be extracted as:
\begin{equation}
    (\mathcal V_i, \mathcal E_i) = (H^{-1}(\mathcal V_c;c_{i}),\mathcal E_c).
    \label{eq:gen_mesh}
\end{equation}
Note that all meshes above share the same connectivity $\mathcal E_0$, so the mesh correspondence is produced. Eq.~\ref{eq:gen_mesh} can be implemented in batch to achieve better efficiency.

%% file: main_text/4exp.tex
\section{Results}

To demonstrate CaDeX as a general and expressive representation, we investigate the performance in modeling three distinct categories: human bodies (Sec.~\ref{sec:exp_dfaust}), animals  (Sec.~\ref{sec:exp_dt4d}) and articulated objects (Sec.~\ref{sec:exp_s2m}). Finally, we examine the effectiveness of our design choice in Sec.~\ref{sec:exp_abl}.

\noindent \textbf{Metrics}: To measure our performance %quantitatively 
for shape and correspondence modeling, we follow the paradigm of~\cite{oflow,lpdc} and use the same metrics: evaluating the reconstruction accuracy using the IoU and Chamfer Distance, and the motion accuracy by correspondence $l_2$-distance error.

\noindent \textbf{Baselines}: 
We compare with the closest model-free dynamic representations. The main baselines described in Sec.~\ref{sec:related} are: \emph{O-Flow}~\cite{oflow} and \emph{LPDC}~\cite{lpdc} for sequence inputs and \emph{A-SDF}~\cite{asdf} for articulated object set inputs.

\subsection{Modeling Dynamic Human Bodies}
\label{sec:exp_dfaust}
\input{./tables/dfaust.tex}

\begin{figure*}[ht!]
    \centering
   \includegraphics[width=\textwidth]{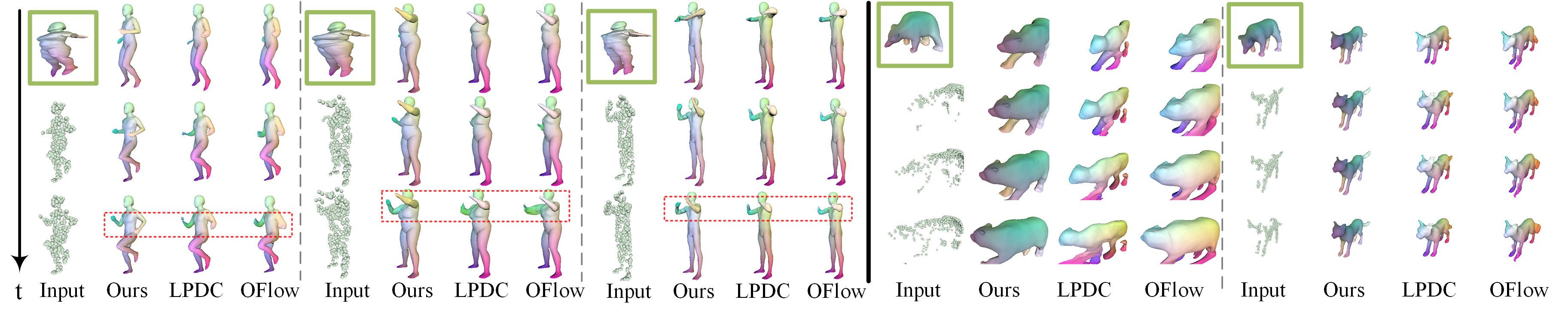}
   \vspace{-2em}
    \caption{
    \textbf{Left}: Human body modeling (Sec.~\ref{sec:exp_dfaust}); \textbf{Right}: Animal body modeling (Sec.~\ref{sec:exp_dt4d}).
    The left top figure marked in the green box is our canonical shape, the first input is not displayed. The colors of the meshes encode the correspondence. More results are in the the Supp.
    }
    \label{fig:dfau_dt4d_results}
\end{figure*}

We first demonstrate the power of modeling the dynamic human body across time. 
We use the same experiment setup, dataset, and split as \cite{oflow,lpdc,lcr}.
The data are generated from D-FAUST~\cite{dfaust:CVPR:2017}, a real 4D human scan dataset. 
Following the setting of \cite{oflow}, the input is a randomly sampled sparse point cloud trajectory (300 points) of 17 frames evenly sampled across time. 
The ground-truth occupancy field as well as the optional surface point correspondence are provided.
Our default model is configured by using the \emph{ST}-encoder (Sec.~\ref{sec:deformation_encoder}) and an NVP homeomorphism. The following tables and sections assume such a configuration if not otherwise specified.
We test the performance of the \emph{PF}-encoder and the NICE homeomorphism variants as well.
The experiments are divided into training without correspondence (Tab.~\ref{tab:dfau1}) and training with correspondence (Tab.~\ref{tab:dfau2}) tracks for fair comparison between methods. The testing set has two difficulty levels: unseen motion and unseen individuals~\cite{oflow}.

Quantitative comparisons in Tab.~\ref{tab:dfau1},~\ref{tab:dfau2} indicate that our method outperforms state-of-the-art methods by a significant margin.
The qualitative comparison in Fig.~\ref{fig:dfau_dt4d_results} shows 
the advantage of our method in capturing fast moving parts and shape details (marked with red).
We attribute such improvements to two main reasons: First, our factorization of the deformation and its implementation provide a strong regularization that other approaches like O-flow~\cite{oflow} can only achieve with an ODE integration. Additionally, supervising a per-frame implicit reconstruction in our model is equivalent to the dense cross-frame reconstruction supervision in \cite{lpdc}. Second, our shape prior is stored in the learned canonical space (marked green in Fig.~\ref{fig:dfau_dt4d_results}), which is relatively stable across different sequences as shown in the figure. 
When training without correspondence (Tab.~\ref{tab:dfau1}), our method can learn the correspondence implicitly and reach a similar reconstruction performance as \cite{lpdc} in Tab.~\ref{tab:dfau2}, which is trained with dense parallel correspondence supervision.
Comparing the different configurations of our method in Tab.~\ref{tab:dfau2}, the NICE~\cite{nice} version has a performance drop since the deformation is strongly regularized to conserve the volume, but this is achieved by freezing half of the capacity (scale freedom). We note that the naive per-frame encoder (\emph{PF}) works better compared to the spatial-temporal encoder~\cite{lpdc} (\emph{ST}). A potential reason is that the per-frame encoder provides a higher canonicalization level since when deciding the deformation code $c_i$, no information from other frames can be considered, so the \emph{PF} encoder might avoid overfitting.

\subsection{Modeling Dynamic Animals}
\label{sec:exp_dt4d}
\input{./tables/dt4d.tex}
We experiment with a more challenging setting: modeling different categories of animals with one model. We generate the same supervision types as Sec.~\ref{sec:exp_dfaust} based on the DeformingThings4D-Animals~\cite{dt4d} dataset (DT4D-A).
We use 17 animal categories and generate 2 types of input observations: Sparse point cloud input as Sec.~\ref{sec:exp_dfaust} as well as the monocular depth video input from a randomly posed static camera.
We assume that the camera view point estimation problem is solved so all partial observations live in one global world frame.
All models are trained across all animal categories.
We refer the reader  to the supplementary material for more details.
Such a setting is more challenging because animals have both large shape and motion variance across categories. Additionally, the models are required to aggregate information across time and hallucinate the missing parts in depth observation inputs.
Quantitative results of both the sparse point cloud input and the depth input in Tab.~\ref{tab:dt4d} as well as the qualitative results in Fig.~\ref{fig:dfau_dt4d_results} indicate that our method outperforms state-of-the-art methods in these challenging settings. In addition to the reasons mentioned in Sec.~\ref{sec:exp_dfaust}, the improvement when predicting from depth observation can be attributed to our design of the canonical observation encoder (Sec.~\ref{sec:g_encoder}) that explicitly aggregates observations in the CaDeX.

\subsection{Modeling Articulated Objects}
\label{sec:exp_s2m}
\input{./tables/s2m}
\begin{figure}[t!]
    \centering
   \includegraphics[width=\columnwidth]{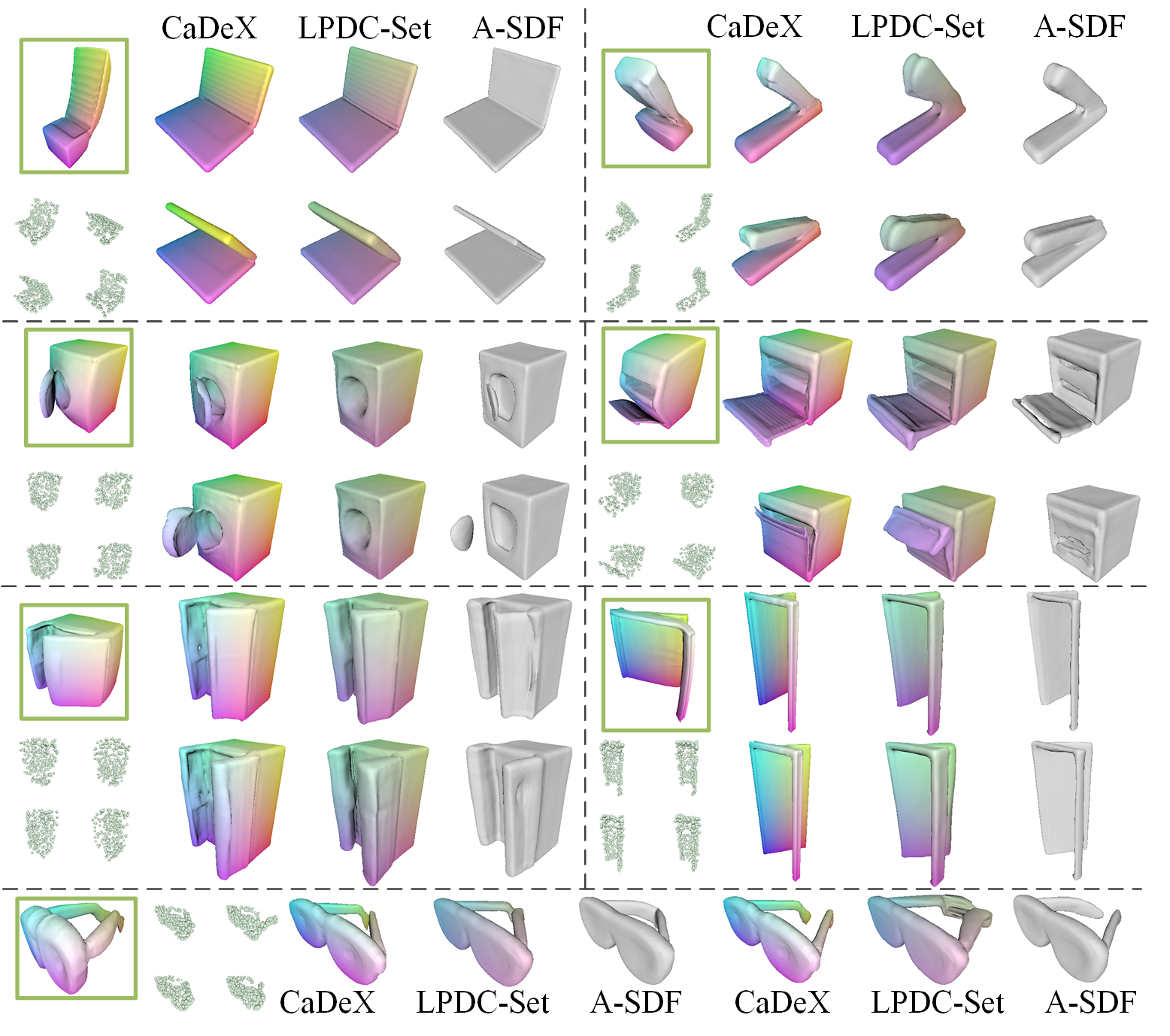}
    \caption{
    Articulated objects modeling (Sec.~\ref{sec:exp_s2m}) with 7 distinct categories. The left-top figure marked in the green box is our canonical shape, the four small figures next to it are the inputs. The first row is the reconstruction of an observed deformation angle and the second is for an unobserved angle. Note that A-SDF has no correspondence so is not colored.
    }
    \label{fig:s2m_results}
\end{figure}
We extend CaDeX from modeling the 4D nonrigid surface sequence to representing semi-nonrigid articulated object sets.
We generate the dataset and inputs as in Sec.~\ref{sec:exp_dt4d} from \cite{asdf} based on Shape2Motion~\cite{wang2019shape2motion}, which contains 7 categories of articulated objects with 1 or 2 deformable angles.
We configure the model with the \emph{SET} encoder (Sec.~\ref{sec:deformation_encoder}) that produces the global dynamic code and then use the articulation angle to query the deformation code for each frame (for details, see the Supplement). During training, we input the sparse point cloud of 4 randomly sampled deformed frames of one object and then use the ground-truth angle to query per-frame deformation codes; finally, the model predicts the occupancy field for 4 seen (input) frames and 4 unseen frames. We supervise both $\mathcal L_R$ and $\mathcal L_C$. 
For completeness, we also predict the articulation angles of the input frames by a small head in the encoder and supervise them. Each category is trained separately for all methods.
Note that A-SDF~\cite{asdf} demonstrates the auto-decoder setup, but it only solves half of our problem without correspondence. Simultaneously solving the shape and the correspondence leads to difficulties when applying an auto-decoder with optimization during testing, so we leave this as a future direction. For fair comparison, we adapt A-SDF with a similar encoder as our model and adapt the decoder to predict the occupancy. We also compare with LPDC~\cite{lpdc} which is also adapted to use a similar encoder as ours.

Tab.~\ref{tab:s2m} summarizes the average performance across 7 object categories while Fig.~\ref{fig:s2m_results} presents the qualitative comparison. 
Both of them show our state-of-the-art performance on modeling general articulated objects.
We produce an accurate reconstruction, while providing the correspondence prediction that A-SDF~\cite{asdf} can not predict. Thus, the marching cube is needed for each frame in \cite{asdf} and results in a longer inference time as shown in Tab.~\ref{tab:s2m}.
Note that our method preserves the topology when deforming the objects (Fig.~\ref{fig:s2m_results} oven) while \cite{asdf} does not have such guarantees. This is the main reason that our method has a performance drop on eyeglasses category since the dataset contains many unrealistic deformations where the legs of the eyeglasses get crossed. % and the topology changes.
Additionally, our method models more details in the moving parts (e.g, the inner side of the refrigerator door in Fig.~\ref{fig:s2m_results}) due to the learned canonical space, which provides a stable container for the shape prior.

\subsection{Ablation Study}
\label{sec:exp_abl}

We show the effectiveness of our design as the following:
First, we replace the invertible canonical map with a one-way MLP that maps the deformed coordinates to the canonical space (such setting is similar to~\cite{yenamandra2021i3dmm,deformedif,zheng2021deep}). Since the mapping is one-way, we supervise the correspondence by enforcing the consistency in the canonical space. Every frame needs a separate application of marching cubes to extract meshes in this version. 
Second, we remove the geometry encoder in the canonical space and obtain the global geometry embedding via a latent fusion using the ST-encoder~\cite{lpdc}. We demonstrate the performance of the deer subcategory from DT4D-A~\cite{dt4d} with point cloud inputs (Sec.~\ref{sec:exp_dt4d}).
Tab.~\ref{tab:abl} shows the performance difference, where we observe a significant performance decrease as well as longer inference times when using MLP instead of homeomorphisms. Additionally, we observe the drop in reconstruction accuracy when removing the geometry encoder in the canonical space (Sec.~\ref{sec:canonical_shape}). We present more details in the supplementary material.

\input{./tables/abl.tex}

%% file: tables/dfaust.tex
\begin{table}[t!]
\centering
\scalebox{0.80}{
\begin{tabular}{@{}cccc|ccc@{}}
\toprule
\multirow{2}{*}{Method}   & \multicolumn{3}{c|}{Seen Individual}              & \multicolumn{3}{c}{Unseen Individual}             \\ \cmidrule(l){2-7} 
                             & IoU↑   & CD↓   & Corr↓ & IoU↑   & CD↓   & Corr↓ \\ \midrule
\multicolumn{1}{c|}{PSGN-4D~\cite{fan2017point}} & -      & 0.108 & 3.234 & -      & 0.127 & 3.041 \\
\multicolumn{1}{c|}{ONet-4D~\cite{onet}} & 77.9\% & 0.084 & -     & 66.6\% & 0.140 & -     \\
\multicolumn{1}{c|}{O-Flow~\cite{oflow}}  & 79.9\% & 0.073 & 0.122 & 69.6\% & 0.095 & 0.149 \\
\multicolumn{1}{c|}{LCR~\cite{lcr}}     & 81.8\% & 0.068 & -     & 68.2\% & 0.100 & -     \\
\multicolumn{1}{c|}{LCR-F~\cite{lcr}}   & 81.5\% & 0.068 & -     & 69.9\% & 0.094 & -     \\
\multicolumn{1}{c|}{Ours} & \textbf{85.5\%} & \textbf{0.056} & \textbf{0.100} & \textbf{75.4\%} & \textbf{0.074} & \textbf{0.126} \\ \bottomrule
\end{tabular}
}
\caption{Results on D-FAUST~\cite{dfaust:CVPR:2017} human bodies, trained \textbf{without} correspondence supervision.}
\label{tab:dfau1}
\end{table}

\begin{table}[t!]
\centering
\scalebox{0.80}{
\begin{tabular}{@{}lccc|ccc@{}}
\toprule
\multicolumn{1}{c}{\multirow{2}{*}{Method}} & \multicolumn{3}{c|}{Seen Individual}              & \multicolumn{3}{c}{Unseen Individual}             \\ \cmidrule(l){2-7} 
\multicolumn{1}{c}{}            & IoU↑   & CD↓   & Corr↓ & IoU↑   & CD↓   & Corr↓ \\ \midrule
\multicolumn{1}{c|}{PSGN-4D~\cite{fan2017point}}    & -      & 0.101 & 0.102 & -      & 0.119 & 0.131 \\
\multicolumn{1}{c|}{O-Flow~~\cite{oflow}}     & 81.5\% & 0.065 & 0.094 & 72.3\% & 0.084 & 0.117 \\
\multicolumn{1}{c|}{LPDC~\cite{lpdc}}       & 84.9\% & 0.055 & 0.080 & 76.2\% & 0.071 & 0.098 \\
\multicolumn{1}{c|}{Ours(NICE)} & 85.4\% & 0.051 & 0.082 & 75.6\% & 0.070 & 0.104 \\
\multicolumn{1}{c|}{Ours(ST)}   & 86.7\% & 0.046 & 0.077 & 78.1\% & 0.063 & 0.095 \\
\multicolumn{1}{c|}{Ours(PF)}               & \textbf{89.1\%} & \textbf{0.039} & \textbf{0.070} & \textbf{80.7\%} & \textbf{0.055} & \textbf{0.087} \\ \bottomrule
\end{tabular}
}
\caption{Results on D-FAUST~\cite{dfaust:CVPR:2017} human bodies, trained with correspondence supervision.}
\label{tab:dfau2}
\end{table}

%% file: tables/dt4d.tex
\begin{table}[]
\scalebox{0.80}{
\begin{tabular}{@{}ccccc|ccc@{}}
\toprule
\multirow{2}{*}{Input} & \multirow{2}{*}{Mehtod}     & \multicolumn{3}{c|}{Seen individual}              & \multicolumn{3}{c}{Unseen individual}             \\ \cmidrule(l){3-8} 
                       &                             & IoU↑             & CD↓             & Corr↓           & IoU↑             & CD↓             & Corr↓           \\ \midrule
\multirow{3}{*}{PCL}   & \multicolumn{1}{c|}{O-Flow~\cite{oflow}} & 70.6\%          & 0.104          & 0.204          & 57.3\%          & 0.175          & 0.285          \\
                       & \multicolumn{1}{c|}{LPDC~\cite{lpdc}}   & 72.4\%          & 0.085          & 0.162          & 59.4\%          & 0.149          & 0.262          \\
                       & \multicolumn{1}{c|}{Ours}   & \textbf{80.3\%} & \textbf{0.061} & \textbf{0.133} & \textbf{64.7\%} & \textbf{0.127} & \textbf{0.239} \\ \midrule
\multirow{3}{*}{Dep}   & \multicolumn{1}{c|}{O-Flow~\cite{oflow}} & 63.0\%          & 0.131          & 0.250          & 49.0\%          & 0.228          & 0.374          \\
                       & \multicolumn{1}{c|}{LPDC~\cite{lpdc}}   & 58.4\%          & 0.160          & 0.249          & 45.8\%          & 0.261          & 0.388          \\
                    & \multicolumn{1}{c|}{Ours}   & \textbf{71.1\%} & \textbf{0.094} & \textbf{0.186} & \textbf{55.7\%} & \textbf{0.175} & \textbf{0.301} \\ \bottomrule % w1.25
\end{tabular}
}
\caption{Results on DeformingThings4D~\cite{dt4d} animal bodies, PCL and Dep correspond to the input types.}
\label{tab:dt4d}
\end{table}

%% file: tables/s2m.tex
\begin{table}[]
\centering
\scalebox{0.85}{
\begin{tabular}{@{}cc|ccc|cc@{}}
\toprule
\multicolumn{1}{l}{Input} & \multicolumn{1}{l|}{Method} & IoU↑ & CD↓ & Corr↓ & t(s) & $\theta$ (deg) \\ \midrule
\multirow{3}{*}{PCL} & A-SDF~\cite{asdf} & 55.2\%          & 0.127          & -              & 3.44 & 3.38 \\
                     & LPDC~\cite{lpdc}  & 49.2\%          & 0.171          & 0.230          & 0.53 & 3.00 \\
                     & Ours  & \textbf{58.9\%} & \textbf{0.118} & \textbf{0.160} & 1.12 & 2.75 \\ \midrule
\multirow{3}{*}{Dep} & A-SDF~\cite{asdf} & 53.9\%          & 0.127          & -              & 3.65 & 5.06 \\
                     & LPDC~\cite{lpdc}  & 46.4\%          & 0.195          & 0.269          & 0.54 & 4.85 \\
                     & Ours  & \textbf{56.4\%} & \textbf{0.116} & \textbf{0.161} & 1.26 & 4.34 \\ \bottomrule
\end{tabular}
}
\caption{Results on Shape2Motion~\cite{wang2019shape2motion,asdf} articulated objects, PCL and Dep correspond to the input types. The average performance across 7 categories is reported, we refer the readers to our supplementary for the full table. $t$ is the surface generation average time and $\theta$ is the average angle prediction error. }
\label{tab:s2m}
\end{table}

%% file: tables/abl.tex
\begin{table}[]
\centering
\scalebox{0.80}{
\begin{tabular}{@{}ccccc@{}}
\toprule
\multicolumn{1}{l}{} & IoU↑   & CD↓   & Corr↓ & t(s) \\ \midrule
Full                 & 66.5\% & 0.128 & 0.223 & 1.8  \\
MLP                  & 61.9\% & 0.161 & 0.303 & 20.5 \\
No G-Enc         & 63.4\% & 0.141 & 0.216 & 1.7  \\ \bottomrule
\end{tabular}
}
\vspace{-0.8em}
\caption{Ablation study, $t$ is the average surface generation time.}
\label{tab:abl}
\end{table}

%% file: main_text/5dis.tex
\vspace{-0.5em}
\section{Limitations}

Our method guarantees several desirable properties and achieves state-of-the-art performance on a wide range of shapes, but still has limitations that need future exploration.
Although we can produce continuous deformation across time if $c(t)$ is continuous, the continuity of $c$ is not guaranteed in the ST-encoder~\cite{lpdc} that we use. 
Therefore, when the input undergoes a large discontinuity, we do observe a trembling in the output of both LPDC~\cite{lpdc} and our method. 
Another issue is that although our method preserves the topology, sometimes the real world deformation also results in topology changes.
Future work can explore how to selectively preserve or alter the topology.
Finally, it is currently nontrivial to adapt our method in an auto-decoder framework~\cite{asdf,deepsdf} since it requires simultaneously optimizing the canonical map (deformation) and the canonical shape during testing,
which future work can explore.

%% file: main_text/6conclusion.tex
\vspace{-0.5em}
\section{Conclusion}
We introduced a novel and general representation for dynamic surface reconstruction and correspondence. Our key insight is the factorization of the deformation by continuous bijective canonical maps through a learned canonical shape. We prove that our representation guarantees cycle consistency and topology preservation, as well as (if desired) volume conservation. 
Extensive experiments on reconstructing humans, animals, and articulated objects demonstrate the effectiveness and versatility of our approach.
We believe that CaDeX enables more possibilities for future research on modeling and learning from our dynamic real world.

\small{\noindent\textbf{Acknowledgement}: The authors appreciate the support of the following grants: ARL MURI W911NF-20-1-0080, NSF TRIPODS 1934960, NSF CPS 2038873, ARL DCIST CRA W911NF-17-2-0181, and ONR N00014-17-1-2093.}